\documentclass[twocolumn,showpacs,superscriptaddress,10pt]{revtex4-1}

\usepackage{amsmath,amssymb,graphicx}

\begin{document}

\title{Accurate and fast identification of minimally prepared bacteria phenotypes using Raman spectroscopy assisted by machine learning}

\author{Benjamin Lundquist Thomsen}
\affiliation{Danish Fundamental Metrology, Kogle All\'e 5, 2970 H{\o}rsholm, Denmark}

\author{Jesper B. Christensen}
\affiliation{Danish Fundamental Metrology, Kogle All\'e 5, 2970 H{\o}rsholm, Denmark}

\author{Olga Rodenko}
\affiliation{Danish Fundamental Metrology, Kogle All\'e 5, 2970 H{\o}rsholm, Denmark}

\author{Iskander Usenov}
\affiliation{Institute of Optics and Atomic Physics, Technische Universität Berlin, Straße des 17. Juni 135, 10623 Berlin, Germany}
\affiliation{Art photonics GmbH, Rudower Ch 46, 12489, Berlin, Germany}

\author{Rasmus Birkholm Grønnemose}
\affiliation{Research Unit of Clinical Microbiology, University of Southern Denmark and Odense University Hospital,
J.B. Winsløws Vej 21.2, 5000 Odense, Denmark}

\author{Thomas Emil Andersen}
\affiliation{Research Unit of Clinical Microbiology, University of Southern Denmark and Odense University Hospital,
J.B. Winsløws Vej 21.2, 5000 Odense, Denmark}

\author{Mikael Lassen*,}
\affiliation{Danish Fundamental Metrology, Kogle All\'e 5, 2970 H{\o}rsholm, Denmark}

%\affil[+]{these authors contributed equally to this work}

%\keywords{Keyword1, Keyword2, Keyword3}

%\begin{abstract}

%\end{abstract}

%\flushbottom

\maketitle


\section*{Abstract}

\textbf{The worldwide increase of antimicrobial resistance (AMR) is a serious threat to human health. To avert the spread of AMR, fast reliable diagnostics tools that facilitate optimal antibiotic stewardship are an unmet need. In this regard, Raman spectroscopy promises rapid label- and culture-free identification and antimicrobial susceptibility testing (AST) in a single step. However, even though many Raman-based bacteria-identification and AST studies have demonstrated impressive results, some shortcomings must be addressed. To bridge the gap between proof-of-concept studies and clinical application, we have developed machine learning techniques in combination with a novel data-augmentation algorithm, for fast identification of minimally prepared bacteria phenotypes and the distinctions of methicillin-resistant (MR) from methicillin-susceptible (MS) bacteria. For this we have implemented a spectral transformer model for hyper-spectral Raman images of bacteria. We show that our model outperforms the standard convolutional neural network models on a multitude of classification problems, both in terms of accuracy and in terms of training time. We attain more than 96$\%$ classification accuracy on a dataset consisting of 15 different classes and 95.6$\%$ classification accuracy for six MR-MS bacteria species. More importantly, our results are obtained using only fast and easy-to-produce training and test data. }    

\begin{figure*}[ht]
\begin{center}
\includegraphics[width=\linewidth]{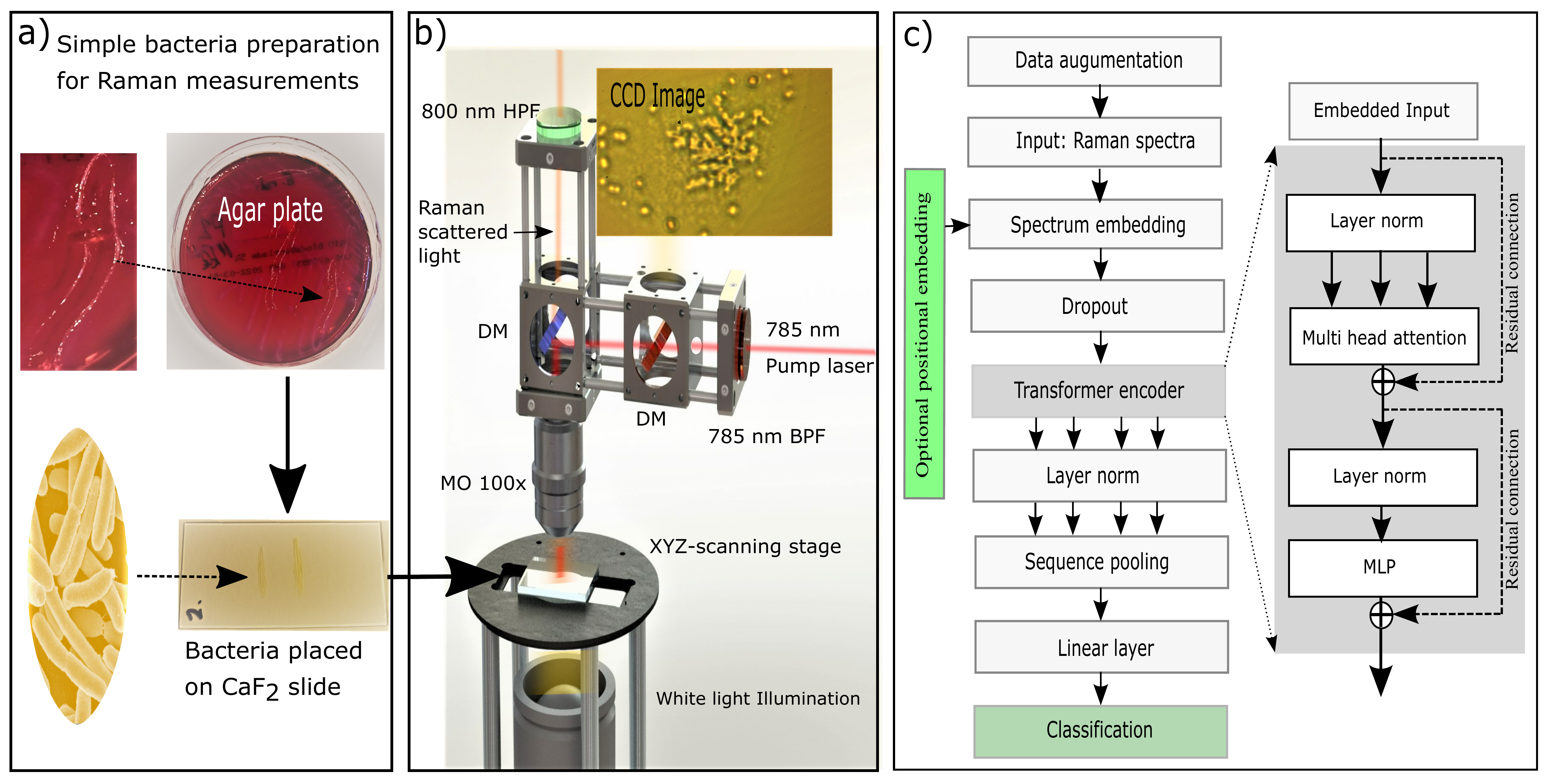} 
\caption{{\textbf{Overview of hardware (Raman microscope) and software (spectral-transformer architecture).}} \textbf{a}) The simple sample preparation of bacteria, where the bacteria from the agar plates are simply transfer directly to the CaF$_2$ objective slides and then measured. The process of transferring and finding the bacteria takes less than one minute. \textbf{b}) Schematic of the home-built Raman microscope. The Raman microscope uses an excitation wavelength of 785 nm,  since it has been found to be optimal for identifying bacteria, as it largely avoids fluorescence, and still gives a sufficiently high Raman signal to enable detection by a CCD at a reasonable signal-to-noise ratio (SNR). A 100$\times$ microscope objective (MO) is used for focusing the excitation laser (spot size $\sim$ 1 $\mu$m), collection of the Raman scattered light, and visual imaging. Raster scanning is achieved with an automated XYZ stage. A dichroic mirror (DM) (high pass 750 nm) is used to couple the visible illumination light to a CCD for imaging and localizing bacteria, while another DM (high pass 805 nm) separates the Raman scattered light from the pump. An additional high pass filter (HPF, 800 nm) and a band pass filter (BPF, 785 nm $\pm$ 10 nm) is used for filtering of the 785 nm pump. The in-build microscope has a field-of-view of approximately 60\,$\mu$m $\times$ 60\,$\mu$m, and Raman spectra are collected at a wavenumber shift of 700-1600 cm$^{-1}$ by a Horiba spectrometer. \textbf{c}) The block diagram of the developed machine learning tool. The spectral transformer (ST) consists of an optional positional embedding layer, followed by a dropout layer. The next layer is a transformer-encoder block that sequentially contains, layer normalisation, multihead attention, layer normalisation, and then a multilayer perceptron (MLP) with a GELU non-linearity. The transformer-encoder output is followed by layer normalisation and a sequence pooling layer. Finally, the output layer is a fully connected linear layer.}
\label{fig:1}
\end{center}
\end{figure*}

\section*{Introduction}

While some health crises, such as the corona pandemic, are unforeseeable and require immediate measures, others are slow to develop, intractable in nature, but may in time become a larger threat to human health \cite{WHO-global-action-plan,UN-Report}. An example of the latter is antimicrobial resistance (AMR) \cite{Stekel2018,NEILL2016,murray2022global,WorldbankAMR}. AMR occurs when microbes, such as bacteria and fungi survive exposure to compounds that would normally inhibit their growth or kill them. This drives a process of selection, allowing strains with resilience to grow and spread. Although AMR is a naturally occurring process, it is dramatically accelerated by selective pressures such as overuse of antimicrobials \cite{tenover2006mechanisms,holmes2016understanding,dadgostar2019antimicrobial,nathan2020resisting,aslam2018antibiotic}. Conventional techniques used for identifying AMR in bacteria are disk diffusion test, epsilometer test, and microdilution, which require culturing and can take days \cite{khan2019current,Reverter2020}. The long processing time of these techniques can be life threatening to the infected patient, but is also problematic, since the pathogenic bacteria might spread and infect more people. Therefore it is common practice to prescribe broad spectrum antibiotics to patients, which leads to unnecessary treatment \cite{Amann2019}. Thus, the already widespread and increasing inadequacy of antimicrobial therapy, is attributed to the overuse of antimicrobials in healthcare and agriculture. \cite{holmes2016understanding,abushaheen2020antimicrobial,murray2022global}. In 2019, the World health organisation (WHO) declared AMR as “one of the 10 biggest global public health threats facing humanity” and according to a report released by the UN ad hoc Interagency Coordinating Group on Antimicrobial Resistance (IACG), if no action is taken, antimicrobial resistant pathogens could annually cause 10 million deaths by 2050 \cite{UN-Report}. 

To mitigate the potential disaster of a post antibiotic era, organisations such as WHO and IACG, are calling for the development of fast point-of-care diagnostic that will facilitate treatment with targeted antimicrobials \cite{WHO-global-action-plan,murray2022global}. To achieve this many different technologies have been studied \cite{barghouthi2011universal,Florio2018,hou2019current,khan2019current,wang2021applications}. One very promising technology is Raman spectroscopy (RS). RS is a technique based on inelastic scattering that occurs when photons collide with molecules and allows for unique signal decomposition for a wide range of molecules \cite{jones2019raman}. Importantly, RS has the advantage of being fast, low-cost, label-free, and does not necessarily require pre-analytical cultivation. Several studies have shown that the capabilities of RS can be significantly strengthened when assisted by chemometric tools and machine learning (ML) \cite{wang2021applications,ralbovsky2020towards,guo2021chemometric,lorenz2017cultivation,novelli2018culture,Ho2019,ashton2011raman,eberhardt2015advantages,chang2019antibiotic,strola2014single,de2021biochemical,Duraipandian2019}. Yet, some shortcomings must be addressed before it will be a viable platform for reliable bacteria identification and point-of-care diagnostics applications. Foremost, RS is sensitive to factors such as the growth stage of the analysed cells, changes in measurement environment and inconsistency in sample preparation \cite{lorenz2017cultivation}. Therefore, it is convenient to prepare samples in a way that reduce the difficulty of classification. Approaches such as preparing single bacteria or monolayer mats of bacteria are unfortunately complex and require expertise, custom equipment, and can take hours \cite{Ho2019,garcia2018label,rousseau2021fast}. Moreover, inconsistencies in sample preparation may cause changes to Raman spectra, necessitating more data for ML models to capture the breadth of variations needed to reach clinically relevant accuracies \cite{wang2021applications}. Furthermore, RS bacteria studies dealing with patient samples are rare and it cannot be assumed that using data from laboratory cultured samples will allow for accurate identification of genuine patient samples. Additionally, there is little to no endorsement of standards for Raman measurement parameters and sample preparation methods- and parameters \cite{lorenz2017cultivation,guo2021chemometric}. This lack, stupendously impedes consolidation of databases, slowing the aggregation of big-data that could be used for clinical applications. To reach clinically relevant accuracies using RS, these issues must be addressed and solving them all will require collective effort.

In this work, we focus on addressing the issues of simple sample preparation and changes in measurement environment \cite{kloss2015toward}. We reduce sample preparation to merely transferring the bacteria to the measurement environment (as shown in Fig.~\ref{fig:1}a), minimising the issue of sample inconsistency. This procedure comes with the additional benefit of removing sample preparation as an inhibiting parameter for data consolidation. Moreover, to alleviate the situation of limited data availability for ML model training, we have developed a novel spectral transformer (ST) ML model that is efficient after training on both small- and large RS bacteria datasets. To feed the ST with good representative training data we have developed a novel data augmentation algorithm, henceforth known as {\texttt{NoiseMix}}. We demonstrate that our ST model in conjunction with {\texttt{NoiseMix}} allows for accurate classification of both single bacteria and multilayer mats of bacteria all in one go, while importantly  only relying on fast- and easy-to-produce training data acquired on thick multilayer mats of bacteria. To our knowledge, this is a completely new approach for acquiring training data and subsequently classification of bacteria using RS assisted by ML. Explicitly, we demonstrate the capabilities of our developed ST ML model and {\texttt{NoiseMix}} on a dataset consisting of 12 classes of bacteria from minimally prepared bacteria samples and 3 non-bacteria classes. We find that {\texttt{NoiseMix}} improves the average classification accuracy by 12.9 $\%$ for the four different tests compared to only utilising class balancing and slope removal. Further, we demonstrate that the ST model can distinguish between antibiotic resistant- and susceptible phenotypes, i.e. MR \textit{S. epidermidis} (MRSE), MS \textit{S. epidermidis} (MSSE), 2 types of MR \textit{S. aureus} (MRSA), and two types of MS \textit{S. aureus} (MSSA). We obtain identification accuracies of 97.7$\%$ and 94.6$\%$ between MRSE-MSSE and MRSA-MSSA isolates, respectively. In addition to identifying minimally prepared samples, we perform detailed benchmark tests of the ST by comparing it with a convolutional neural network (CNN) developed in the work by Ho.~et al.~on multiple RS bacteria datasets \cite{Ho2019}. We find that our ST model significantly outperforms the CNN model in terms of computation time, which is improved by one order of magnitude, and that it generally outperforms the CNN model in terms of classification accuracy, for which we achieve an improvement of 7.5 $\%$ compared to the reference CNN model \cite{Ho2019}.

\section*{Results}

A home-built Raman microscope is used to acquire training and validation datasets of minimally prepared bacteria samples. The schematics of our Raman microscope for acquiring Raman hyper-spectral maps is shown in Fig.~\ref{fig:1}b). The reason for using a home-built system is that it gives us the possibility to optimize the Raman microscopes signal-to-noise ratio (SNR) and tailor the system to the task of detecting bacteria. Hereby we can acquired Raman spectra using very short measurement times down to 0.1 second and also have a relatively cheap system compared to commercial Raman microscopes. For more details about the microscope and spectrometer, see the Methods section.

\subsection*{Fast- and simple training data acquisition}
Successful classification of bacteria using RS and ML relies heavily on having a large training database to be used in the model training- and validation steps. The collection of data therefore often becomes as important as ML algorithms themselves, since over- or underrepresented data will lead to biased predictions. If RS is to be considered for fast in-situ diagnostic applications, the complexity and time cost of sample preparation must be significantly reduced \cite{kloss2015toward,franco2019advances,pahlow2015isolation}. To explore how much we can simplify and reduce sample preparation time- and complexity, we experimented with simply transferring bacteria samples from a bacterial monoculture directly to a CaF$_2$ objective slide followed by Raman raster-scan measurements. This approach causes the depth of the bacterial samples to naturally vary from mono- to multilayer-deep mats, causing large variations in the intra-sample SNR \cite{garcia2018label}. Training data maps produced in this manner necessitates manual segmentation, as the maps may contain areas without bacteria (background). To avoid the need for manual segmentation, we instead produce training data exclusively from measurements of multilayer bacterial mats. However, data originating from measurements of multilayer bacterial mats has a limited SNR distribution compared to data acquired from bacterial mono- to multilayers. With the purpose of synthetically recreating the natural variances that may appear in test data, we produce training data by varying the spectroscope integration time from 0.1 to 1 seconds (10 averages for each acquisition). With this process, and an automated Raman spectroscopy setup (see Methods), we acquire several thousands of training spectra a day. Our final reference bacteria database contains in excess of 5200 raw Raman spectra for each of the 12 bacterial species and 3 non-bacteria species. All raw data is linearly pre-processed by a simple procedure (see Methods) before being used for either data augmentation, model training or model prediction.

\subsection*{Data augmentation using {\texttt{NoiseMix}}}
With inspiration from computer vision wherein “extra” training data is often augmented for example by rotating, flipping, blurring, or adding white noise to images, we have developed a data augmentation algorithm ({\texttt{NoiseMix}}) that allows us to synthetically create additional training data and thereby enhance model generalization and performance. The {\texttt{NoiseMix}} augmentation algorithm (see Supplementary Material for technical details), works by taking fast and easy to produce Raman spectra from multilayer bacterial mats, and then mixing in data with even more “noise” of both the measurement surface/environment, and noise data from measurements in the environment. In addition to increasing the quantity of training-data examples, {\texttt{NoiseMix}}, as implemented here, brings two further advantages. Firstly, it allows a synthetic extension of the RS dataset towards the region of lower SNR distributions. In this sense, training data with an arbitrarily low SNR can in principle be realized, although the SNR is in practice kept above a certain minimum value to avoid inclusion of training examples consisting of pure noise. Remarkably, we find that the {\texttt{NoiseMix}} augmentation algorithm allows high-accuracy identification of single bacteria although the original training examples are exclusively gathered from multilayer bacterial mats. Secondly, the {\texttt{NoiseMix}} algorithm provides a means to leverage all data of class-imbalanced datasets by ensuring that all classes are represented by the same amount of data in each training epoch.

\begin{figure*}[th]
\begin{center}
    \includegraphics[width=1\linewidth]{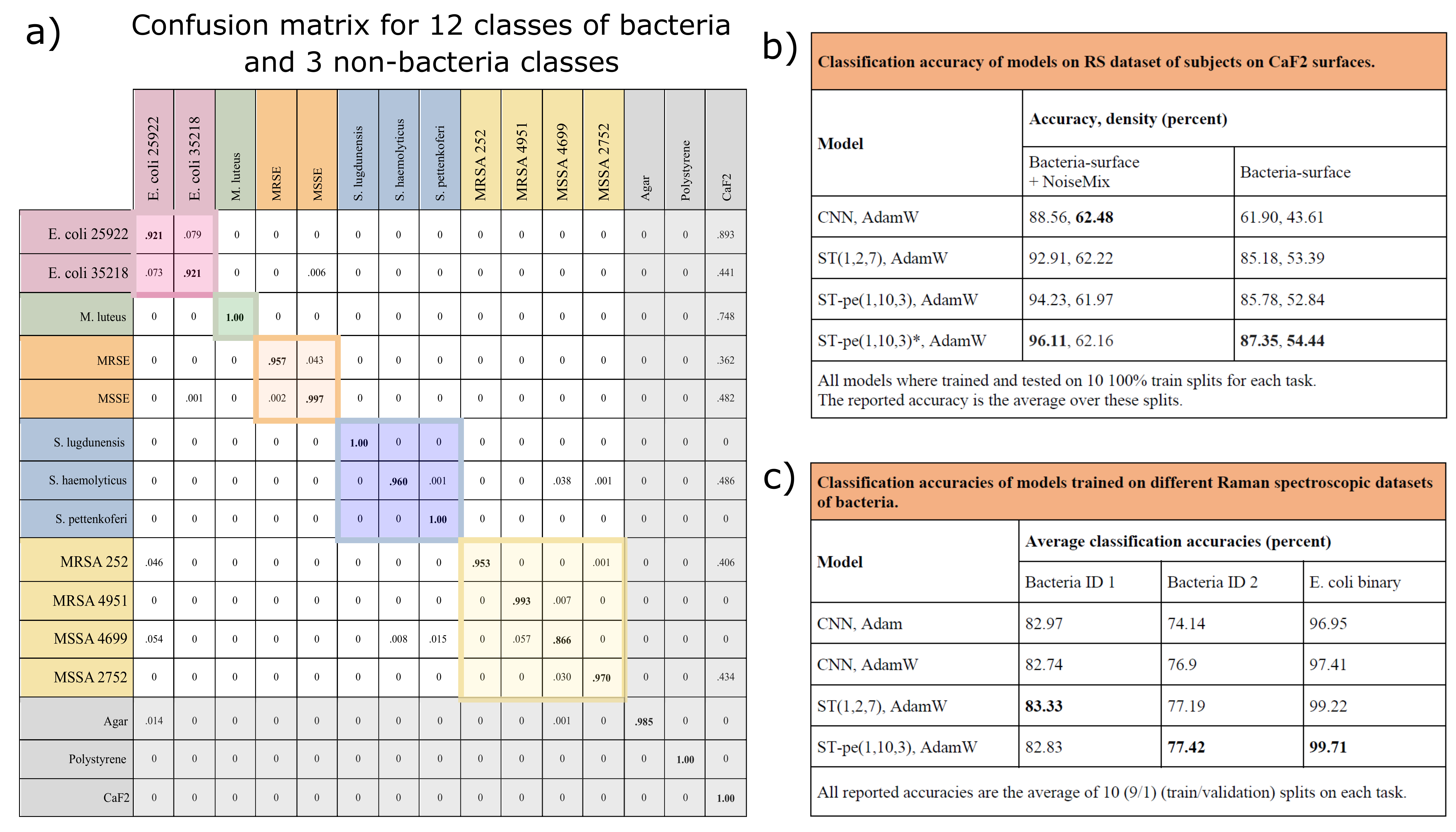}
    \caption{{\textbf{Performance overview of bacteria identification with the ST model and \texttt{NoiseMix} algorithm.}} {\textbf{a)}} Shows the confusion matrix obtained for the classification task including 12 bacterial- and 3 non-bacterial classes (agar, polysterene, and CaF$_2$). The CaF$_2$-classification column (to the right) contain non-zero elements as the sample surfaces were in some cases only partly covered by bacteria. For this reason, the non-bacterial classes are greyed out as they are not included in the bacterial identification accuracy.  {\textbf{b)}} Shows a comparison in performance between four different ML models trained both with and without applying {\texttt{NoiseMix}}. The results displayed in the confusion matrix are obtained using the ST-pe(1,10,3)* model trained using a batch size of 300 and the AdamW optimizer. The other three models are also trained using the AdamW optizimer but with a smaller batch size of 100. The model accuracies (and densities) represent averages over 10 training splits. In {\textbf{c)}} we show the results of a benchmark test between the CNN- and the ST models when applied to three different classification tasks. The three datasets are described in the Supplementary Material. In this case the reported accuracies represent the average of 10 runs using a training/validation split of 90$\%$/10$\%$.}
    \label{fig:2}
\end{center}
\end{figure*}

\subsection*{Raman identification of bacteria with the spectral transformer}
Bacteria identification using RS has in recent years experienced a significant performance gain as deep-learning techniques such residual connections and CNNs have proved more capable than more classical supervised learning methods such as logistic regression and support vector machines \cite{Ho2019, lussier2020deep, lu2020combination}. To improve even further upon this, we have developed an attention-based deep-learning model inspired by current state-of-the-art in computer vision and natural language programming. The ST model (sketched in Fig.\,\ref{fig:1}(c) and explained in more detail in Methods) is a compact version of the standard transformer encoder \cite{parmar2018image}, but differs by using sequence pooling to map the sequential outputs of the transformer to a singular class.

Our ST model architecture is initially parameterized by three arguments ST(-pe)($i$,$j$,$k$), where $i$ is the depth of the transformer encoder, $j$ is the number of heads in the multi-head attention layer, $k$ is the multilayer perceptron ratio, and the inclusion of -pe signifies an optional positional embedding. The three arguments were treated as additional hyperparameters of our model and were selected using a Tree-structured Parzen Estimator, using one training and validation split on the isolate classification task \cite{Ho2019}, i.e.~we did not use our own RS data to fit our model architecture to the task at hand.

Our main results are summarized in Fig.\,\ref{fig:2}, with a) displaying the confusion matrix of the 15-class (12 bacteria and 3 non-bacteria) classification task. An overall accuracy in excess of 96$\%$ is achieved over the 12 bacteria classes using an ST-pe(1,10,3) model trained using the AdamW optimizer and applying {\texttt{NoiseMix}}. Fig.~\ref{fig:2}b) breaks down a comparison in accuracy between multiple different ML models, with and without applying {\texttt{NoiseMix}}, on the same 15-class classification task. We observe that augmenting training data with {\texttt{NoiseMix}} significantly improves model performance in the test phase for both the three ST models and the reference CNN model, and we find that both ST model architectures outperform the reference CNN model on our 15-class dataset.

In addition to model accuracy (given as ratio of correct bacteria classifications to total number of bacteria classifications) we also report a density metric in Fig.\,\ref{fig:2}b). The density (or bacteria coverage) is defined as the ratio of bacteria classifications to the total amount of classifications made in each test. This metric is included in our case because part of our test data for some bacteria consists of data from background (see e.g.~Fig.~\ref{fig:3} below), and hence not every measurement should be affiliated with a bacteria type. Notably, the density metric is significantly increased by applying {\texttt{NoiseMix}} which is attributed to the algorithms' ability to improve classification for low-SNR signals.

Fig.~\ref{fig:2}c) compares model classification performance on three different bacteria datasets (for an overview of the datasets and the applied training process, see Supplementary Material). Datasets "Bacteria ID 1", and "Bacteria ID 2" originate from the work of Ho et.\,al.\,\cite{Ho2019}. For these datasets we observe only a marginal improvement, on average, by using either of the two tested ST models. The final dataset "E.\,coli binary" originates from our own RS database and contains Raman spectra from \textit{E.\,coli} ATCC25922 and \textit{E.\,coli} ATCC35218. For this dataset, the ST models again significantly outperforms the CNN models, suggesting that the ST architecture may perform well on a broader task of spectroscopy-based classification problems.

As a final performance benchmark, we compared the computation time of the ST model with that of the reference CNN model \cite{Ho2019} (see Supplementary Material). We generally observe an approximate speed-up of one order of magnitude in favor of the developed ST model. However, it should be noted that a small amount of this speed-up may be caused by differences in model hyperparameters, such as weight decay, parameter amount, and learning rate, and that the difference therefore cannot solely be attributed to the model architectures.

\subsection*{Visualization and differentiation of {\textit{E.~coli}} isolates }
\begin{figure*}[t]
\begin{center}
    \includegraphics[width=1\linewidth]{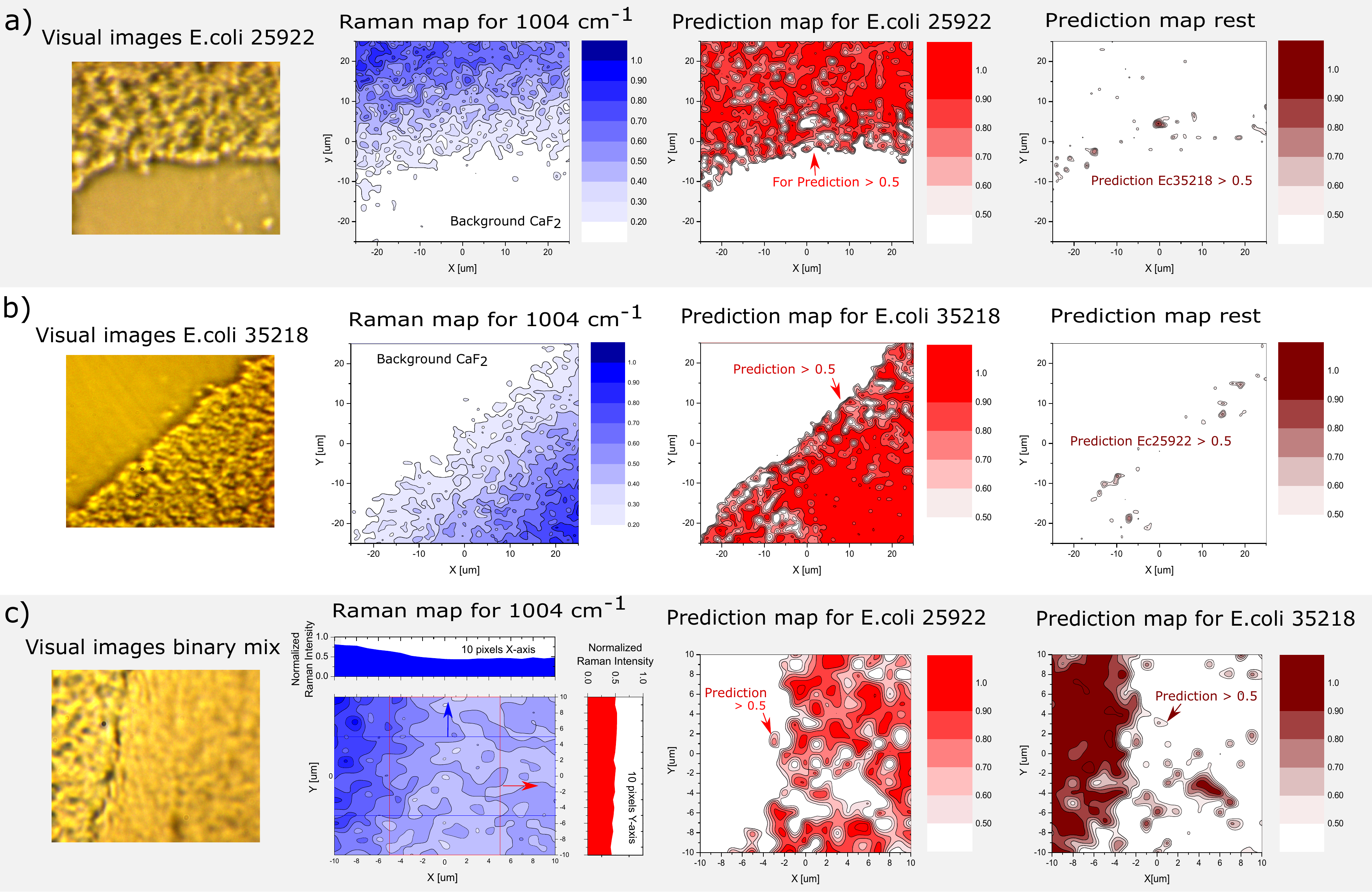}
    \caption{{\textbf{Raman imaging and ST identification of \textit{E.\,coli} ATCC25922 and \textit{E.\,coli} ATCC35218.}} The first column shows visual images of the measurement areas, and illustrates bacterial depths ranging from single- to multiple layers (4-6 $\mu$m thick). Raman maps are shown in the second column for a Raman shift of 1004 cm$^{-1}$, assigned to the ring breathing mode vibrations of l-Phe, and finally ST prediction maps are shown in the third and fourth columns. The size of the maps are 51\,$\mu$m $\times$ 51\,$\mu$m and they each consist of 2601 Raman spectra (700-1600 cm$^{-1}$) with 1 $\mu$m spacing between the points. Raman spectra are acquired with 10-times averaging of 0.5 s integration. \textbf{a)} Raman measurements of \textit{E.\,coli} ATCC25922. The overall prediction rate (density surface coverage) is 49.1\% for \textit{E.\,coli} ATCC25922, 10.4\% for \textit{E.\,coli} ATCC35218 and 40.2\% for CaF$_2$ background. For the rest of the bacteria/classes the total prediction rate sums to 0.3\%. The prediction map to the right shows prediction for the rest of the classes plotted for >0.5, where only \textit{E.\,coli} ATCC35218 has values higher than 0.5. \textbf{b)} Measurements of \textit{E.\,coli} ATCC35218. The overall prediction rate is 8.0\% for \textit{E.\,coli} ATCC25922, 49.0\% for \textit{E.\,coli} ATCC35218 and 42.8\% for background. For the rest of the bacteria/classes the prediction sums to 0.2\%. Again the ST makes a few \textit{E.\,coli} ATCC25922 misclassifications.  \textbf{c)} Raman measurements for a binary mixture of \textit{E.\,coli} ATCC25922 and \textit{E.\,coli} ATCC35218 resulting in a prediction rate (surface coverage) of 48.8\% and 51.2\%, respectively. The ST does in this case not make any misclassifications. All prediction of other bacteria than the two \textit{E.\,coli} is zero. For all three acquired maps the ST prediction maps agrees very well with the Raman map and the visual map.}
    \label{fig:3}
\end{center}
\end{figure*}

For better understanding of the capability and performance of our developed ST model and {\texttt{NoiseMix}} we visualize the analysis by showing the Raman maps and ST prediction maps. We conduct tests both on monocultures and mixes of monocultures as seen in Fig.~\ref{fig:3}. Fig.~\ref{fig:3}a) and b) show visual images of the test area for two monocultures of \textit{E.\,coli} ATCC25922 and \textit{E.\,coli} ATCC35218, respectively. The Raman maps are acquired with a step size of 1 $\mu$m over an area 50$\mu$m x 50$\mu$m and are plotted for the ringbreathing mode vibrations of l-Phe (Raman shift 1004 cm$^{-1}$). Each Raman map consist of 2601 points and each point (Raman spectrum, 700-1600 cm$^{-1}$) is acquired from 10 averages with an integration time of 0.5 seconds, with a complete measurement time of 217 minutes. Comparing the visual images, Raman intensity maps, and the prediction maps in Fig.~\ref{fig:3}a) and b), we find an excellent agreement between the different forms of visualization. From the Raman intensity contour maps depicted in Fig.~\ref{fig:3}, it is evident that the Raman intensity decreases in the demarcation zone between CaF$_2$ and bacteria. This is in part due to a thinner bacteria layer (monolayer) and in part due to the smaller laser-bacteria overlap. Without the {\texttt{NoiseMix}} method the ST prediction maps would underestimate the region with bacteria coverage and make significant more misclassifications in the demarcation zone between the CaF$_2$ and bacteria. Thus, the resulting decrease in SNR of the Raman signals has the consequence that the ML models, which is exclusively trained on multilayer bacterial mats underestimates the region covered with bacteria, and make a large number of misclassifications in the demarcation zone. However, by applying {\texttt{NoiseMix}} in the training phase, the ST model becomes extremely efficient even at detecting and identifying low concentrations of bacteria (monolayers) even though the original training data only contains measurements of multilayer bacterial mats. Which is attributed to the {\texttt{NoiseMix}} algorithms' ability to improve classification for low-SNR Raman signals. We define a accuracy for a class as: correct/(crosses + correct), where crosses are all wrong predictions with values above >0.5 and excluding prediction of background (CaF$_2$). This gives a accuracy of 87.3\% and 87.9\% for Fig.~\ref{fig:3}a) and Fig.~\ref{fig:3}b), respectively. Comparing the accuracies with the surface coverage we find that our ST classifier for this specific case is undetermined in approximately 10\% of the time, where the prediction rate is lower than 0.5. The 15-class ST classifier makes primarily the misclassifications in the demarcation zone. Note that by increasing the integration time to 2 seconds, or more, this would decrease the occurrence of misclassifications, but has the consequence that the completely measurement time of one Raman map with 2601 Raman spectra would take more than 14 hours.

Fig.~\ref{fig:3}c) shows a random mix of \textit{E.\,coli} ATCC25922 and \textit{E.\,coli} ATCC35218 cultures. The two monoculture samples are transferred directly to the CaF$_2$ objective slide, where they are mixed and subsequently measured. From the visual image and the Raman map, no information about the mixture of \textit{E.\,coli} ATCC35218 and \textit{E.\,coli} ATCC25922 can be obtained. The only information that is deduced is that the layer is slightly thicker in the left side, which can be seen from the 10 pixels projection of the contour plot onto the $x$- and $y$-axis. However, from the ST prediction map we clearly see the mixture of the two \textit{E.\,coli} bacteria. We find that the ST model, with {\texttt{NoiseMix}} applied in the model training phase, did not make any misclassifications and predicted only correct species, namely the \textit{E.\,coli}, with an estimated density ratio of 48.8\% of \textit{E.\,coli} ATCC25922 and 51.2\% of \textit{E.\,coli} ATCC35218. The reason for this impressive classification result, where only \textit{E.\,coli} is predicted is due to the thick layer of bacteria distribution of 4-6 $\mu$m, Thus the Raman signal SNR is always relatively high. Further, we find an overall accuracy of 98.1\% for \textit{E.\,coli} ATCC25922 and \textit{E.\,coli} ATCC35218, where the last 1.9\% are undetermined data points with an equal prediction rate of 0.5, which sums up to approximately 49 points in the Raman map.

\begin{figure*}[t]
\begin{center}
    \includegraphics[width=1\linewidth]{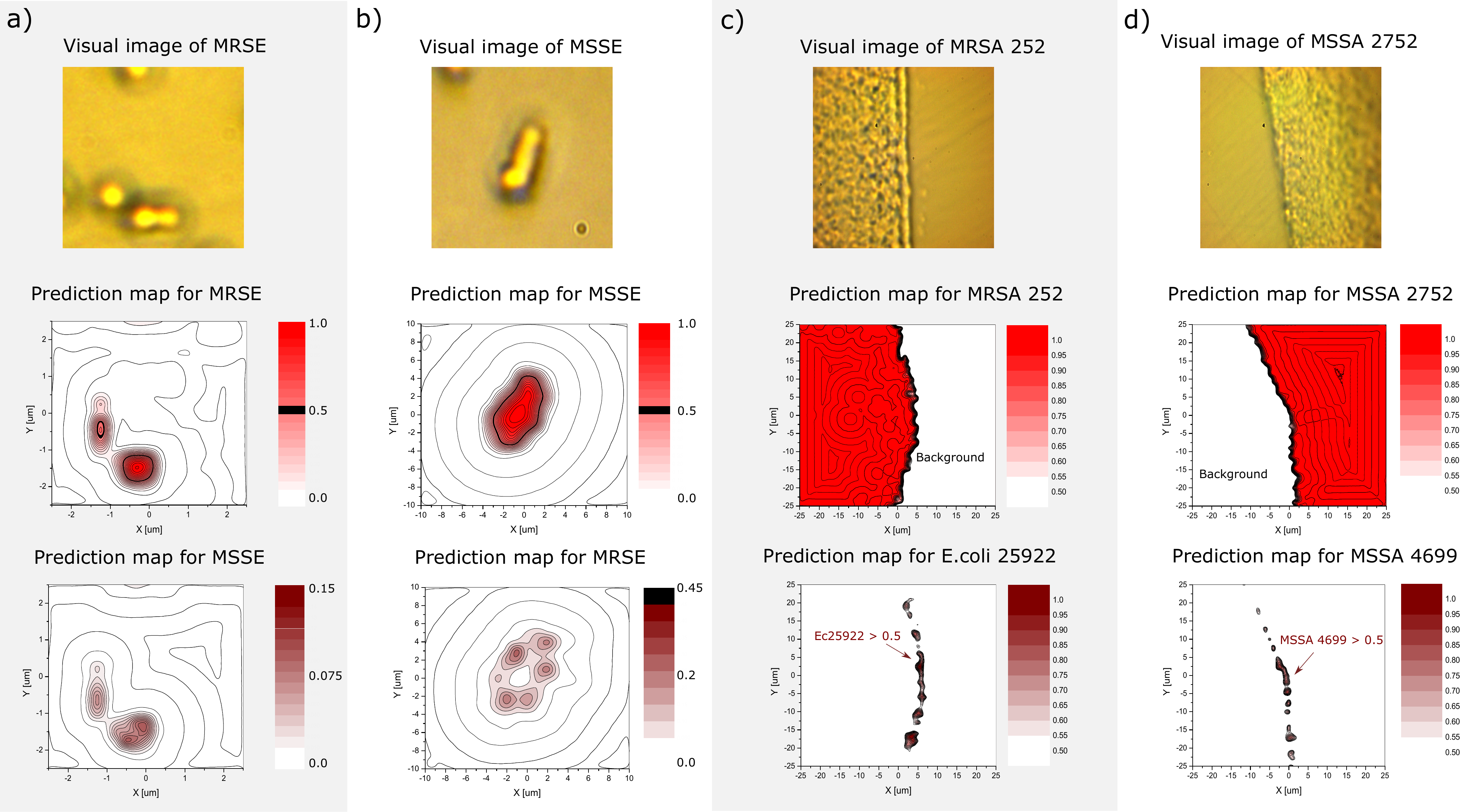}
    \caption{{\textbf{Raman measurements and differentiation of antibiotic resistant phenotypes.}} The figure shows the visual images and the ST prediction maps for a) Methicillin-resistant \textit{S. epidermidis} ATCC35984 (MRSE), b) Methicillin-sensitive \textit{S. epidermidis} ATCC14990 (MSSE), c) Methicillin-resistant \textit{S. aureus} MRSA ATCC252, and d) Methicillin-sensitive \textit{S. aureus} MSSA ATCC2752. The bacteria distribution ranges from a single bacterium to thick layers (4-6 $\mu$m thickness) of bacteria. From the visual images we see that a) MRSE and b) MSSE are acquired for single (few) bacterium.  The integration time used was 10 second for acquiring each Raman spectra and average 10 times. For MRSE the size of the maps are 5\,$\mu$m $\times$ 5\,$\mu$m and consist of individual 441 Raman spectra (700-1600 cm$^{-1}$) with 0.25 $\mu$m spacing between the points. For MSSE the size of the maps are 10\,$\mu$m $\times$ 10\,$\mu$m with 1 $\mu$m spacing between the points and consist of 441 individual Raman spectra. The integration time used was 2 second for acquiring each Raman spectra and average 10 times. In both cases the ST makes no misclassifications, however there is small certainty for the bacteria to be MSSE and MRSE as seen in the MSSE and MRSE prediction maps in a) and b), respectively. In c) and d) the visual and prediction maps for MRSA and MSSA are shown. The  50\,$\mu$m $\times$ 50\,$\mu$m and consist of 2601 Raman spectra (700-1600 cm$^{-1}$) with 1 $\mu$m spacing between the points. The integration time used is 0.5 second and average 10 times for acquiring each spectra.}
    \label{fig:4}
\end{center}
\end{figure*}

\subsection*{Testing for antibiotic resistance and -susceptibility }
Fig.~\ref{fig:4} shows measurements and test for the differentiation of antibiotic resistant bacteria. For this proof-of-concept AST we collect Raman maps from clinical isolates of MR $\textit{S. epidermidis}$ ATCC35984 (MRSE), MR $\textit{S. aureus}$ ATCC252 (MRSA 252), MR $\textit{S. aureus}$ ATTCC4951 (MRSA4951) and on MS $\textit{S. epidermidis}$ ATCC14990 (MSSE), MS $\textit{S. aureus}$ ATTCC4699 (MSSA 4699), and MS $\textit{S. aureus}$ ATCC2752 (MSSA 2752). The overall model performance of the 15-class classifier on the MR-MS classification task can be seen in the confusion matrix in Fig.~\ref{fig:2}. The ST classifier also contains \textit{S. lugdunensis}, \textit{S. haemolyticus} and \textit{S. pettenkoferi} these strains were chosen to represent biological variation, potential cross-interference for making a more difficult classification task for the ST, and to create a realistic view on the possibilities of our technique. 
Notably, we find that the ST distinguishes between MRSE and MSSE isolates of \textit{S.\,epidermidis} with a prediction accuracy in excess of 99.5$\%$. Examples of predictions maps for the MRSE, MSSE, MRSA 252, and MSSA 2752 and reference bacteria can be seen in Fig.~\ref{fig:4}. In Fig.~\ref{fig:4}c) and Fig.~\ref{fig:4}d) the measurement of MRSA and MSSA is shown for two monocultures of MRSA 252 and MSSA 2752 reference bacteria, respectively. Fig.~\ref{fig:4}c) shows that the ST estimate the prediction rates (density surface coverage) to be 40.5$\%$ for CaF$_2$ background, 56$\%$ for MRSA 252, 0.4$\%$ for MSSA 2752 and 3.1$\%$ for \textit{E.\,coli} ATCC25922. Again, it is evident that in the demarcation zone between CaF$_2$ and the MRSA bacteria the misclassification rates are higher, due the decrease in SNR. For this measurement the ST does indeed make a 69 misclassifications, which can be seen from Fig.~\ref{fig:4}c), where predictions rate between 0.5 and up to 0.99 for \textit{E.\,coli} ATCC25922 is found. However, this could also be related to contamination of the test sample. In Fig.~\ref{fig:4}b) measurements of MSSA 2752 are shown. We find that the prediction rates (surface coverage) are 41.6$\%$ for CaF$_2$ background, 55.4$\%$ for MSSA 2752 and 3$\%$ for MSSA 4699. The ST has a few misclassification, where the ST is predicting the bacteria to be MSSA 4699, as seen in Fig.~\ref{fig:4}b), again these are mostly found in the demarcation zone and is therefore related to the low SNR found here. By increasing the integration time, to 2 seconds or more, would have circumvented these misclassifications, however the since the map consisting of 2601 individual spectra, the acquisition time would take more than 14 hours. From the confusion matrix we find that the overall performance of the 15 class ST classifier has a prediction accuracy of 94.6$\%$, for the sub-matrix of the two MRSA and two MSSA isolates. If we compare our results with a binary classifier used in ref.~\cite{Ho2019}, where they distinguish between MRSA and MSSA with an achieving 89.1$\%$ accuracy, we find that our ST model clearly outperforms the CNN model. Note that if measurements only are conducted on thick layers of monocultures of bacteria we find that the ST has a very high accuracy. Not shown visually, but we find as an example for MSSA 2752 and MRSA 4951 accuracies of 99.7\% and 99.9\%, respectively. Which might not be surprising since the training validation datasets are very similar.

In addition to distinguishing between antibiotic resistant and antibiotic sensitive isolates, we also test our developed ST and {\texttt{NoiseMix}} method on single bacteria (few bacteria), as can be seen in Fig.~\ref{fig:4}a) and b). The maps are acquired with 10 second integration-time, however without {\texttt{NoiseMix}} we found that the ST model could not identify any bacteria, thus demonstrating how {\texttt{NoiseMix}} improves the sensitivity of the ML models. The prediction rates (density surface coverage) for Fig.~\ref{fig:4}a) are 96.8$\%$ CaF$_2$ background, 2.9$\%$ MRSE and 0.3$\%$ MSSE. The highest prediction peak for MSSE is only 0.15. Thus, the ST does not make any misclassification bewteen MRSE and MSSE or any other bacteria class. For Fig.~\ref{fig:4}b) we find that the prediction rates are 93$\%$ CaF$_2$ background, 0.2$\%$ \textit{E.\,coli} ATCC35218, 1.3$\%$ MRSE, and 5.5$\%$ of MSSE. Again, the ST does not make any misclassification between MRSE and MSSE, since the highest prediction peak found for MSSE are 0.45.  We it remarkably that our ST together with {\texttt{NoiseMix}} also allows high-accuracy identification of single bacteria although the original training examples are exclusively gathered from multilayer bacterial mats.

\begin{figure*}[ht]
    \begin{center}
    \includegraphics[width=1\linewidth]{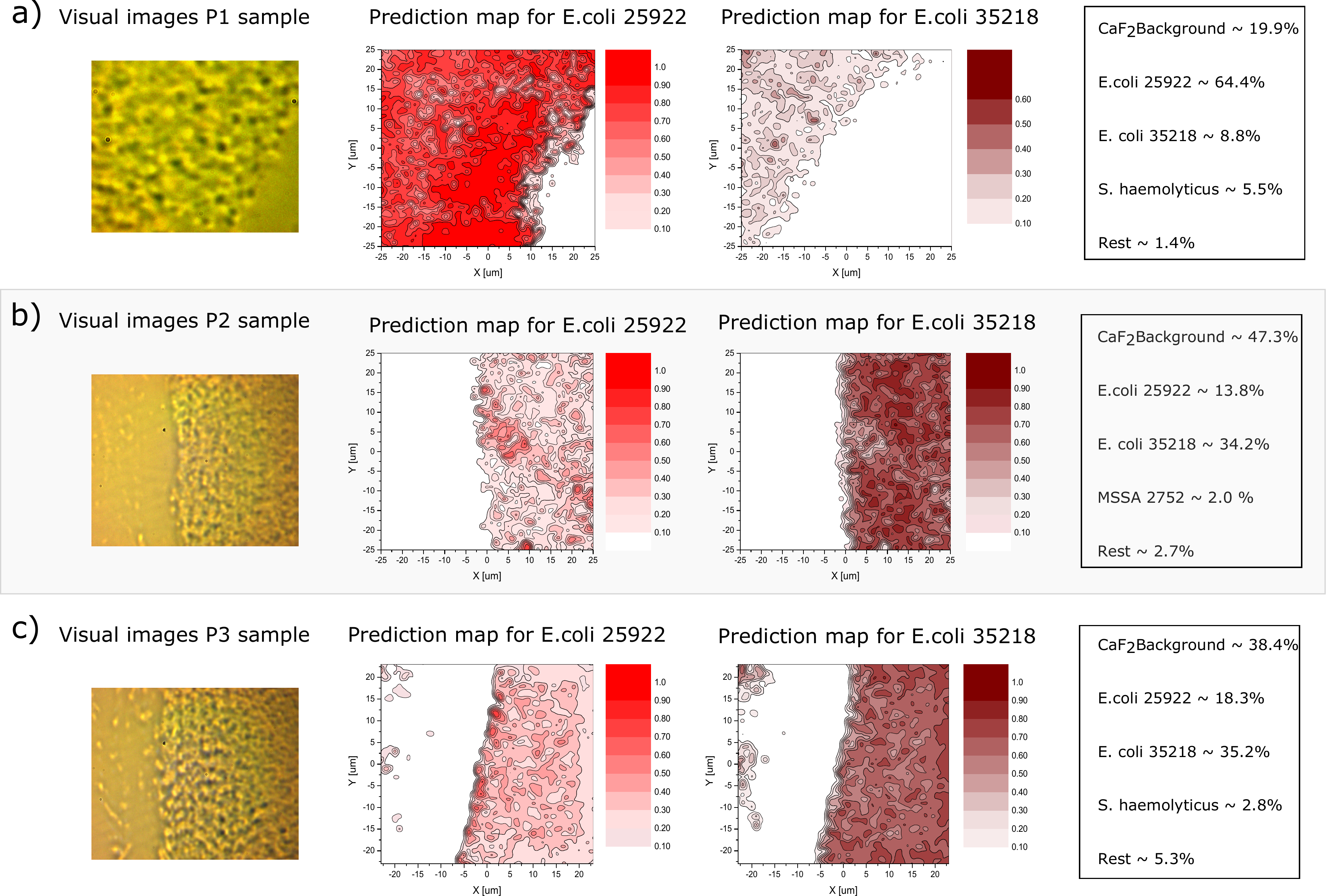}
    \caption{{\textbf{Raman measurements and ST classifications of three cultured \textit{E.\,coli} patient samples.}} The figure shows the visual images of the measurement areas, where it can be seen that the bacteria distribution ranges again from a deep layer (4-6 $\mu$m thick) to single bacteria depth and the ST prediction maps for \textit{E.\,coli} ATCC25922 and \textit{E.\,coli} ATCC35218. The size of the maps are 50\,$\mu$m $\times$ 50\,$\mu$m and they each consist of 2601 Raman spectra (700-1600 cm$^{-1}$) with 1 $\mu$m spacing between the points. The integration time used is 0.5 second for acquiring the spectra and average 10 times per point/spectra. The table shows the overall prediction rates for CaF$_2$ background, \textit{E.\,coli} ATCC25922, \textit{E.\,coli} ATCC35218 and the rest of the classes. Specific we see that a) Patient sample 1 has a overall prediction rate for the other bacteria of 6.9\%, b) Patient sample 2 of 4.7\% and c) patient sample 3 of 8.1\%. However, the accuracies (prediction rate >0.5) are for P1: 98.5\%, P2: 99.4\% and P3: 98\% that the sample is \textit{E.\,coli}.  }
    \label{fig:5}
    \end{center}
\end{figure*}

\subsection*{Identification of clinical {\textit{E.~coli}} isolates}
In Fig.~\ref{fig:3} we investigated the performance of the ST and {\texttt{NoiseMix}} on \textit{E.\,coli} reference bacteria from the same clinical monoculture isolates. However, in order to demonstrate that our ST also potentially works for clinical patient isolates we conduct tests on three random \textit{E.\,coli} clinical patient isolates obtained from Odense University Hopitals clinical microbiology laboratory. The \textit{E.\,coli} isolates P1, P2, and P3 (shown in Fig.~\ref{fig:5}) are isolated from urine and were species identified from indole spot test (positive) and from plating on CHROMID® CPS ELITE agar plates (Biomérieux, USA). Hereby, making a proof-of-concept of microbial phenotype testing of the 3 patient samples. Note that the ST has never seen these Raman spectra before, since the patient samples have or might have a slightly different phenotype, than the \textit{E.\,coli} reference bacteria used for the training of the ST. We therefore would expect that the ST will return predictions for a mix of the two \textit{E.\,coli} reference bacteria. The visual images and prediction maps for the 3 \textit{E.\,coli} patient samples are shown in Fig.~\ref{fig:5}. From the ST prediction maps we can estimate the overlap (prediction rates) with \textit{E.\,coli} ATCC25922 and \textit{E.\,coli} ATCC35218. We find that the average misclassification for the 3 patient samples are 1.4\% and is partly due to the fact that the ST has not seen any training data for the 3 patient samples before. We again see that the misclassification is mostly found in the demarcation zone between the CaF$_2$ background and the bacteria mat and is therefore also related to the low Raman SNR. A resistance patterns for the three clinical isolates and for the two \textit{E.\,coli} ATCC strains were identified at the Odense University Hospital Clinical Microbiology Laboratory by measuring inhibition zone diameter using the disk diffusion test. From the analysis this could suggest that P1 has indeed the highest similarity with \textit{E.\,coli} ATCC25922. While P2 and P3 shows similar resistance patterns as \textit{E.\,coli} ATCC35218. However, more samples and measurements needs to be conducted in order to verify this and make a conclusive conclusion. However, we can conclude that our ST indeed can distinguish within a few seconds/minutes microbial phenotypes of \textit{E.\,coli} with an average classification accuracy of 98.6$\%$ for the three patient samples.

\section*{Discussion}
For rapid identification of bacteria and to combat the spread of AMR we have conducted a proof-of-concept experiment using RS assisted by ML. We have demonstrated that RS is a promising technology for microbiology studies. For this we have developed an attention-based ML model and a novel data augumentation algorithm ({\texttt{NoiseMix}}) to obtain state-of-the-art results within bacteria identification. The ST model architecture used in this work, is inspired by the success of the visual transformer (VIT) \cite{shao2021adversarial} and compact convolutional transformers (CCT) for their ability to generalize well, when trained on small datasets \cite{Hassani2021}. In contrast to VIT's and CCT's, we found that when dealing with RS data, both splitting the Raman spectra into patches and implementing convolutions to induce an inductive bias, is detrimental to model performance. Furthermore, we have found that limiting model depth increases the efficacy of the model substantially, at least on problems with limited availability of data. We suspect this is due to the capacity of deep transformer models to overfit, which becomes a limiting factor, when intra-sample variance is high, as we observe for our datasets. Which also would be the case for practical implementations of RS for in-situ measurements in clinics and hospitals. We firmly believe that our method of novel data augumentation and RS assisted by our developed ST may close the gap between basic research and practical application in clinical laboratories \cite{kong2015raman}. We explicit demonstrated that our ST outperforms a state-of-the-art domain-specific residual CNN both in terms of accuracy, and computation time \cite{Ho2019}. The significant reduction in computation time importantly reduces both diagnosis time, and the cost of the diagnosis apparatus as the inference time of the ST is fast, even on low cost hardware. The ST models used in this work could also be applied to other spectroscopy-based classification problems such as cancer detection or mineral identification. Our Raman system assisted by the ST model distinguishes between 15 different classes with more than 96$\%$ overall classification accuracy, while the CNN has slightly lower overall classification accuracy 88.6\%. Since this was a proof-of-concept our dataset only contains 15 classes, however, the database can easily be expanded to contain any number of bacteria and non-bacteria.

Comparing our method to methods which are currently being used at hospitals, namely labor and time demanding testing in laboratories, RS assisted by ML is an improvement with respect to speed, coverage, price and handling. Other technologies such as flow cytrometry, polymerase chain reaction and MALDI-TOF mass spectrometry, are also intensively studied for their potential as fast and reliable diagnostic technologies \cite{barghouthi2011universal,Florio2018,hou2019current,khan2019current}. The disadvantage of these technologies is that they require large expensive equipment, need special trained personnel and they cannot be used locally as a point-of-care diagnostic/screening tool. Importantly, mass spectrometers require cultivation, has difficulties to discriminate closely related bacterial species and to differentiate some antibiotic resistance phenotypes, such as MRSA and MSSA \cite{wang2021applications}. In contrast, we demonstrate that our RS assisted by the ST and {\texttt{NoiseMix}} approach enable accurate classification of different bacteria phenotypes, namely \textit{E.\,coli}, \textit{S. Epidermidis}, and \textit{S. Aureus}. Importantly, our result is obtained with easy to produce Raman training data that was gathered from deep monoculture mats of bacteria. With this simple preparation approach for acquiring training data, we consistently achieve diagnosis times of less than a few minutes, if culturing is disregarded. The significance of our data collection method is paramount, as our approach facilitates easy, fast and cheap development of big datasets, which is crucial for clinical application. Consequently, it is possible to simply create training data from cultured bacteria, and then embed background and contaminant noise with {\texttt{NoiseMix}} into the fast-and easy-to-produce training data. This would allow for both fast data production, fast sample preparation, and would not need any form of filtering or culturing of the bacteria. It is therefore reasonable to assume that our approach can readily be adopted for direct diagnosis of sepsis from genuine patient samples, without any prepossessing. Assuming this, accurate diagnosis and therefore treatment with targeted antimicrobial can be achieved within few minutes.

\section*{Methods}

\subsection*{Sample preparation}
The bacteria come from bacterial isolates which were cultured overnight on agar plates and were sealed with parafilm and stored at 5$^{\circ}$C until sample preparation. Storage time varied, but was not found to result in spectral changes to strain or phenotype characteristics. All other sample preparation conditions were kept consistent between samples. Test samples were prepared separately from samples used for training, to ensure classification was not influenced by differences in sample preparation. To prepare samples for Raman measurement, a sample was simply transferred from a single colony directly to a sterilized CaF$_2$ Raman-grade objective slide.

\subsection*{Datasets and training details}

Bacteria-surface + {\texttt{NoiseMix}} and Bacteria-surface: The Bacteria-surface training dataset is made up of three integration times for each class. The dataset consists of 12 classes of bacteria (E. coli ATCC35218, E. coli ATCC25922, methicillin-resistant S. epidermidis ATCC35984
(MRSE), methicillin-sensitive S. epidermidis ATCC14990 (MSSE), Micrococcus luteus, S. lugdunensis, S. haemolyticus, S. pettenkoferi, methicillin-resistant S. aureus ATCC252, methicillin-resistant S. aureus ATTCC4951, methicillin-sensitive S. aureus ATTCC4699, methicillin-sensitive S. aureus ATCC2752, and 3 non-bacteria classes, calcium fluoride, (CaF$_2$), agar, and polystyrene beads. The data of the bacterial classes in the Bacteria-surface training dataset were acquired by measuring over CaF$_2$ slides, which were completely covered by multilayer bacterial mats. The data of the CaF$_2$ background class in the Bacteria-surface training dataset, was acquired by measuring clean CaF$_2$ slides. The data of the agar class in the Bacteria-surface training dataset, was acquired by measuring over CaF$_2$ slides covered by a deep layer of agar. The data of the polystyrene class in the Bacteria-surface training dataset, was acquired by measuring over CaF$_2$ slides, which were completely covered by polystyrene beads. For tests using NoiseMix e.g. in Fig.~\ref{fig:2}, Fig.~\ref{fig:3} and Fig.~\ref{fig:4}, the CaF$_2$ and agar Bacteria-surface training data, is used as mixing inputs for the algorithm. The Bacteria-surface test dataset used in Fig.~\ref{fig:2}, consists of 12 classes of bacteria and 3 non-bacteria classes. Each class in the Bacteria-surface test dataset is represented by one measurement over a partially covered CaF$_2$ surface. The bacteria classes in the Bacteria-surface test dataset, is therefore not represented by the same number of bacteria Raman spectra. The Bacteria-surface validation dataset is produced in the same way as the Bacteria-surface test dataset but does not contain all 15 classes. The measurements shown in Fig.~\ref{fig:3}, Fig.~\ref{fig:4} and Fig.~\ref{fig:5} are acquired following the same procedure used to produce the Bacteria-surface test dataset. Pre-processing of the Bacteria-surface training dataset data, consists of normalising each spectrum between 0-1.
Pre-processing of data shown in Fig.~\ref{fig:3}, Fig.~\ref{fig:4} and Fig.~\ref{fig:5} of the Bacteria-surface test and validation data, consists of two steps. ($i$) the slope of the spectra’s are removed by subtracting the linear function between the start and end values of the spectra’s, and ($ii$) a normalisation step in which each Raman spectra is normalised between 0 and 1. For the results shown in Fig.~\ref{fig:3}, Fig.~\ref{fig:4} and Fig.~\ref{fig:5}, we use 100$\%$ of data from the Bacteria-surface training dataset for training and then use the held-out Bacteria-surface validation dataset for model selection. As the validation set is produced with the same procedure as the actual test dataset it is a better indicator of model classification efficacy.

\textbf{Bacteria ID 1:}
The models are trained on the reference dataset from Stanford \cite{Ho2019}, which consists of 30 bacterial and yeast isolates with 2000 spectra for each of the 30 isolates. The models were then fine-tuned on the reference fine-tuning dataset which consists of 30 bacterial and yeast isolates with 100 spectra for each of the 30 isolates \cite{Ho2019}. The models are subsequently tested on the reference test dataset, consisting of 30 bacterial and yeast isolates with 100 spectra for each of the 30 isolates \cite{Ho2019}.

\textbf{Bacteria ID 2:} The models were trained only on the reference fine-tuning dataset, and subsequently tested on the reference test dataset \cite{Ho2019}.

\textbf{\textit{E.\,coli} binary:} The models were trained and tested on binary datasets consisting of \textit{E.\,coli} ATCC35218 and \textit{E.\,coli} ATCC25922. The data of the \textit{E.\,coli} binary datasets were acquired by measuring over CaF$_2$ slides, which were covered by multilayer bacterial mats. The \textit{E.\,coli} binary training dataset has 5180 spectra for each class, and each class is made up of two different integration times, each containing 2590 spectra. The \textit{E.\,coli} binary test dataset has 2590 spectra for each class, and the integration times are different from those of the training set. Pre-processing for the \textit{E.\,coli} binary datasets, consists of two steps performed automatically without user intervention: ($i$) a baseline-correction step using Zhangfit \cite {Zhang2010}, and ($ii$) a normalisation step in which each Raman spectra is normalised between 0 and 1.

\subsection*{Raman microscope}
The Raman microscope for acquiring Raman data is shown in Fig.~\ref{fig:1}b). The Raman microscope uses a 785-nm excitation laser (TA pro, Toptica, Germany) with 60 mW of power. The pump beam is spatial cleaned with a 1 meter long single-mode (SM) fiber (PANDA PM FC/PC to FC/APC Patch Cable) with 5.3 $\mu$m mode-field diameter. A long-working-distance $100\times$ microscope objective (MO) (LMPLN-IR/LCPLN-IR, numerical aperture NA = 0.85) from Olympus is used both for imaging, focusing the excitation laser and collecting the backscattered light. The bacteria samples are placed on Raman graded calcium fluoride (CaF$_2$) objective slides and the position is controlled with an automated XYZ scanning stage. A dichroic mirror (DM) (high pass 750 nm, Semrock) is used to couple the visible illumination light to a charge-coupled device (CCD) for imaging. A second DM (high pass 800 nm) is used to separate the Raman signal from the pump. Additional filters (high pass, 800 nm, Semrock) and (band pass, 785 nm $\pm$ 10 nm, Semrock) are used for filtering of the 785-nm pump. A 5 m long multi-mode (MM) fiber (Ø200 µm, 0.39 NA, FC/PC to FC/PC Patch Cables) collects the Raman signal and directs it to the spectrometer. For acquiring Raman spectra, we use a HR320 Horiba spectrometer. All measurements were performed with a slit size of 300 $\mu$m and the grating used has a line density of 950 L/mm. A thermoelectrically cooled charge-coupled device (CCD) is used for detection (Synapse, 1024 × 256 with each pixel size of 26 $\mu$m). The CCD pixels are binned in clusters of 2x20 pixels to reduce noise and hereby increase SNR. With each acquired Raman spectra consisting of 480 points in the range from 700-1600 cm$^{-1}$, the spectral resolution of the spectrometer is approximately 10\,cm$^{-1}$.

\subsection*{Raman maps}
To control the position and change the sampling point for RS, we use an XYZ scanning stage from Applied Scientific Instrumentation (ASI). The ASI stepper motors provide precise control through the use of closed-loop DC servomotors employing high‐resolution encoders for positioning and feedback. The XY stage has a range of travel of 100~mm $\times$ 100~mm and a positional accuracy of approximately 200 nm. Custom made Python software was developed for the automation of the complete Raman microscope to asynchronously control the scanning stage and Horiba spectrometer for acquiring hyperspectral Raman maps of the bacteria samples.

\subsection*{Spectroscopic calibration}
For spectral calibration (and optimization) of the Raman microscope and and calibration of the translation stage, we use polystyrene beads ranging in size from 1-5\,$\mu$m. The polystyrene beads are comparable in size to bacteria and constitute multiple Raman peaks in the same Raman shift region as the bacteria. From the measurements and ST prediction maps we estimate that the spatial resolution of the Raman maps are $\approx$ 2 $\mu$m $\pm 500$ nm) and for the ST prediction map are $\approx$ 3 $\mu$m $\pm 500$ nm). 

\subsection*{Pre-processing of data}
The raw Raman spectra were initially cleaned from cosmetic spikes. Subsequently the linear function between the start and end values of each spectrum is identified and subtracted. As a final pre-processing step, the spectra were individually normalized to the range between zero and one. Notably, we have also investigated baseline-correction methods using Zhangfit [36], however we found that any kind of non-linear baseline removal was detrimental to model performance, especially when used in conjunction with {\texttt{NoiseMix}}.

\subsection*{\texttt{NoiseMix} and the data-augmentation process}
To improve model performance in the test phase, we apply data augmentation in the model traning phase. The algorithm {\texttt{NoiseMix}} works by randomly selecting and subsequently mixing bacteria spectra $S_{bacteria}(\nu)$ and background spectra $S_{bg}(\nu)$. An augmented Raman spectra $S_{bacteria}^{(aug)}(\nu)$ is then given by
\begin{equation*}
   S_{bacteria}^{(aug)}(\nu) = (1 - \alpha) S_{bacteria}(\nu) +\alpha S_{bg}(\nu)  
\end{equation*}
where $\alpha$ is chosen randomly from a uniform distribution in the range $[0, \alpha_{max}]$, and $\alpha_{max} <1$ is an upper bound for the contribution of background spectra.

\subsection*{Spectral-transformer architecture}
The ST ML model developed here is a compact version of the standard transformer encoder \cite{parmar2018image}, but differs in that it uses sequence pooling to map the sequential outputs to a singular class  The structure of the ST model can be seen in Fig.~\ref{fig:1}c). It consists of an optional positional embedding layer (ST-pe), followed by a dropout layer. The next layer is a block that sequentially contains, layer norm, multihead attention (MHA), layer norm, and then a multilayer perceptron (MLP) with a GELU nonlinearity. This is followed by layer norm, and then a sequence pooling layer. Finally, the output layer is a fully connected linear layer. Our ST architecture is parameterised by three arguments ST($i$,$j$,$k$), where $i$ is the depth of the transformer encoder, $j$ is the number of heads in the MHA layer, and $k$ is the multilayer perceptron ratio. Hence, in the ST(1,2,7) version, the transformer encoder has a depth of 1, the MHA layer has 2 heads, and the hidden layer dimension of the MLP is 7 times larger than the MLP input dimension. These hyperparameters, as well as all hyperparameters used for training, were selected using a Tree-structured Parzen Estimator, using one training and validation split on the isolate classification task \cite{Ho2019}.

\subsection*{Accuracy and density}
As we have included non-bacteria background classes in our model, we opted to use two performance metrics: accuracy and density.
Accuracy is defined in the usual sense as the ratio of correct bacteria classifications to the total number of bacteria classifications. Density on the other hand is a measure of bacteria coverage, and is given as the number of bacteria classifications to the total number of classifications.

\bibliographystyle{naturemag}
\bibliography{sample.bib}

\begin{thebibliography}{10}
\expandafter\ifx\csname url\endcsname\relax
  \def\url#1{\texttt{#1}}\fi
\expandafter\ifx\csname urlprefix\endcsname\relax\def\urlprefix{URL }\fi
\providecommand{\bibinfo}[2]{#2}
\providecommand{\eprint}[2][]{\url{#2}}

\bibitem{WHO-global-action-plan}
\bibinfo{author}{{World Health Organization}}.
\newblock \bibinfo{title}{Global action plan on antimicrobial resistance}
  (\bibinfo{year}{2015}).

\bibitem{UN-Report}
\bibinfo{author}{on~Antimicrobial~Resistance, I. C.~G.}
\newblock \bibinfo{title}{Report to the secretary-general of the united
  nations}  (\bibinfo{year}{2019}).

\bibitem{Stekel2018}
\bibinfo{author}{Stekel, D.}
\newblock \bibinfo{title}{First report of antimicrobial resistance pre-dates
  penicillin}.
\newblock \emph{\bibinfo{journal}{Nature}} \textbf{\bibinfo{volume}{562}}
  (\bibinfo{year}{2018}).

\bibitem{NEILL2016}
\bibinfo{author}{O'Neill, J.}
\newblock \bibinfo{title}{Tackling drug-resistant infections globally: final
  report and recommendations} (\bibinfo{year}{2016}).

\bibitem{murray2022global}
\bibinfo{author}{Murray, C.~J.} \emph{et~al.}
\newblock \bibinfo{title}{Global burden of bacterial antimicrobial resistance
  in 2019: a systematic analysis}.
\newblock \emph{\bibinfo{journal}{The Lancet}}  (\bibinfo{year}{2022}).

\bibitem{WorldbankAMR}
\bibinfo{author}{bank group, W.}
\newblock \bibinfo{title}{Drug-resistant infections: A threat to our economic
  future}  (\bibinfo{year}{2017}).

\bibitem{tenover2006mechanisms}
\bibinfo{author}{Tenover, F.~C.}
\newblock \bibinfo{title}{Mechanisms of antimicrobial resistance in bacteria}.
\newblock \emph{\bibinfo{journal}{The American journal of medicine}}
  \textbf{\bibinfo{volume}{119}}, \bibinfo{pages}{S3--S10}
  (\bibinfo{year}{2006}).

\bibitem{holmes2016understanding}
\bibinfo{author}{Holmes, A.~H.} \emph{et~al.}
\newblock \bibinfo{title}{Understanding the mechanisms and drivers of
  antimicrobial resistance}.
\newblock \emph{\bibinfo{journal}{The Lancet}} \textbf{\bibinfo{volume}{387}},
  \bibinfo{pages}{176--187} (\bibinfo{year}{2016}).

\bibitem{dadgostar2019antimicrobial}
\bibinfo{author}{Dadgostar, P.}
\newblock \bibinfo{title}{Antimicrobial resistance: implications and costs}.
\newblock \emph{\bibinfo{journal}{Infection and drug resistance}}
  \textbf{\bibinfo{volume}{12}}, \bibinfo{pages}{3903} (\bibinfo{year}{2019}).

\bibitem{nathan2020resisting}
\bibinfo{author}{Nathan, C.}
\newblock \bibinfo{title}{Resisting antimicrobial resistance}.
\newblock \emph{\bibinfo{journal}{Nature Reviews Microbiology}}
  \textbf{\bibinfo{volume}{18}}, \bibinfo{pages}{259--260}
  (\bibinfo{year}{2020}).

\bibitem{aslam2018antibiotic}
\bibinfo{author}{Aslam, B.} \emph{et~al.}
\newblock \bibinfo{title}{Antibiotic resistance: a rundown of a global crisis}.
\newblock \emph{\bibinfo{journal}{Infection and drug resistance}}
  \textbf{\bibinfo{volume}{11}}, \bibinfo{pages}{1645} (\bibinfo{year}{2018}).

\bibitem{khan2019current}
\bibinfo{author}{Khan, Z.~A.}, \bibinfo{author}{Siddiqui, M.~F.} \&
  \bibinfo{author}{Park, S.}
\newblock \bibinfo{title}{Current and emerging methods of antibiotic
  susceptibility testing}.
\newblock \emph{\bibinfo{journal}{Diagnostics}} \textbf{\bibinfo{volume}{9}},
  \bibinfo{pages}{49} (\bibinfo{year}{2019}).

\bibitem{Reverter2020}
\bibinfo{author}{Reverter, M.} \emph{et~al.}
\newblock \bibinfo{title}{Aquaculture at the crossroads of global warming and
  antimicrobial resistance}.
\newblock \emph{\bibinfo{journal}{Nature communications}}
  \textbf{\bibinfo{volume}{11}}, \bibinfo{pages}{1870} (\bibinfo{year}{2020}).
\newblock
  \urlprefix\url{https://www.nature.com/articles/s41467-020-15735-6#citeas}.

\bibitem{Amann2019}
\bibinfo{author}{Amann, S.}, \bibinfo{author}{Neef, K.} \&
  \bibinfo{author}{Kohl, S.}
\newblock \bibinfo{title}{Antimicrobial resistance (amr)}.
\newblock \emph{\bibinfo{journal}{European Journal of Hospital Pharmacy:
  Science and Practice}} \textbf{\bibinfo{volume}{26}},
  \bibinfo{pages}{175--177} (\bibinfo{year}{2019}).
\newblock \urlprefix\url{https://ejhp.bmj.com/content/26/3/175}.
\newblock \eprint{https://ejhp.bmj.com/content/26/3/175.full.pdf}.

\bibitem{abushaheen2020antimicrobial}
\bibinfo{author}{Abushaheen, M.~A.} \emph{et~al.}
\newblock \bibinfo{title}{Antimicrobial resistance, mechanisms and its clinical
  significance}.
\newblock \emph{\bibinfo{journal}{Disease-a-Month}}
  \textbf{\bibinfo{volume}{66}}, \bibinfo{pages}{100971}
  (\bibinfo{year}{2020}).

\bibitem{barghouthi2011universal}
\bibinfo{author}{Barghouthi, S.~A.}
\newblock \bibinfo{title}{A universal method for the identification of bacteria
  based on general pcr primers}.
\newblock \emph{\bibinfo{journal}{Indian journal of microbiology}}
  \textbf{\bibinfo{volume}{51}}, \bibinfo{pages}{430--444}
  (\bibinfo{year}{2011}).

\bibitem{Florio2018}
\bibinfo{author}{Florio, W.}, \bibinfo{author}{Tavanti, A.},
  \bibinfo{author}{Barnini, S.}, \bibinfo{author}{Ghelardi, E.} \&
  \bibinfo{author}{Lupetti, A.}
\newblock \bibinfo{title}{Recent advances and ongoing challenges in the
  diagnosis of microbial infections by maldi-tof mass spectrometry}.
\newblock \emph{\bibinfo{journal}{Frontiers in Microbiology}}
  \textbf{\bibinfo{volume}{9}}, \bibinfo{pages}{1097} (\bibinfo{year}{2018}).
\newblock
  \urlprefix\url{https://www.frontiersin.org/article/10.3389/fmicb.2018.01097}.

\bibitem{hou2019current}
\bibinfo{author}{Hou, T.-Y.}, \bibinfo{author}{Chiang-Ni, C.} \&
  \bibinfo{author}{Teng, S.-H.}
\newblock \bibinfo{title}{Current status of maldi-tof mass spectrometry in
  clinical microbiology}.
\newblock \emph{\bibinfo{journal}{Journal of food and drug analysis}}
  \textbf{\bibinfo{volume}{27}}, \bibinfo{pages}{404--414}
  (\bibinfo{year}{2019}).

\bibitem{wang2021applications}
\bibinfo{author}{Wang, L.} \emph{et~al.}
\newblock \bibinfo{title}{Applications of raman spectroscopy in bacterial
  infections: principles, advantages, and shortcomings}.
\newblock \emph{\bibinfo{journal}{Frontiers in Microbiology}}
  \textbf{\bibinfo{volume}{12}} (\bibinfo{year}{2021}).

\bibitem{jones2019raman}
\bibinfo{author}{Jones, R.~R.}, \bibinfo{author}{Hooper, D.~C.},
  \bibinfo{author}{Zhang, L.}, \bibinfo{author}{Wolverson, D.} \&
  \bibinfo{author}{Valev, V.~K.}
\newblock \bibinfo{title}{Raman techniques: fundamentals and frontiers}.
\newblock \emph{\bibinfo{journal}{Nanoscale research letters}}
  \textbf{\bibinfo{volume}{14}}, \bibinfo{pages}{1--34} (\bibinfo{year}{2019}).

\bibitem{ralbovsky2020towards}
\bibinfo{author}{Ralbovsky, N.~M.} \& \bibinfo{author}{Lednev, I.~K.}
\newblock \bibinfo{title}{Towards development of a novel universal medical
  diagnostic method: Raman spectroscopy and machine learning}.
\newblock \emph{\bibinfo{journal}{Chemical Society Reviews}}
  \textbf{\bibinfo{volume}{49}}, \bibinfo{pages}{7428--7453}
  (\bibinfo{year}{2020}).

\bibitem{guo2021chemometric}
\bibinfo{author}{Guo, S.}, \bibinfo{author}{Popp, J.} \&
  \bibinfo{author}{Bocklitz, T.}
\newblock \bibinfo{title}{Chemometric analysis in raman spectroscopy from
  experimental design to machine learning--based modeling}.
\newblock \emph{\bibinfo{journal}{Nature protocols}}
  \textbf{\bibinfo{volume}{16}}, \bibinfo{pages}{5426--5459}
  (\bibinfo{year}{2021}).

\bibitem{lorenz2017cultivation}
\bibinfo{author}{Lorenz, B.}, \bibinfo{author}{Wichmann, C.},
  \bibinfo{author}{St{\"o}ckel, S.}, \bibinfo{author}{R{\"o}sch, P.} \&
  \bibinfo{author}{Popp, J.}
\newblock \bibinfo{title}{Cultivation-free raman spectroscopic investigations
  of bacteria}.
\newblock \emph{\bibinfo{journal}{Trends in microbiology}}
  \textbf{\bibinfo{volume}{25}}, \bibinfo{pages}{413--424}
  (\bibinfo{year}{2017}).

\bibitem{novelli2018culture}
\bibinfo{author}{Novelli-Rousseau, A.} \emph{et~al.}
\newblock \bibinfo{title}{Culture-free antibiotic-susceptibility determination
  from single-bacterium raman spectra}.
\newblock \emph{\bibinfo{journal}{Scientific reports}}
  \textbf{\bibinfo{volume}{8}}, \bibinfo{pages}{1--12} (\bibinfo{year}{2018}).

\bibitem{Ho2019}
\bibinfo{author}{Ho, C.}, \bibinfo{author}{Jean, N.} \& \bibinfo{author}{Hogan,
  C. e.~a.}
\newblock \bibinfo{title}{Rapid identification of pathogenic bacteria using
  raman spectroscopy and deep learning}.
\newblock \emph{\bibinfo{journal}{Nat Commun.}} \textbf{\bibinfo{volume}{10}},
  \bibinfo{pages}{4927} (\bibinfo{year}{2019}).

\bibitem{ashton2011raman}
\bibinfo{author}{Ashton, L.}, \bibinfo{author}{Lau, K.},
  \bibinfo{author}{Winder, C.~L.} \& \bibinfo{author}{Goodacre, R.}
\newblock \bibinfo{title}{Raman spectroscopy: lighting up the future of
  microbial identification}.
\newblock \emph{\bibinfo{journal}{Future microbiology}}
  \textbf{\bibinfo{volume}{6}}, \bibinfo{pages}{991--997}
  (\bibinfo{year}{2011}).

\bibitem{eberhardt2015advantages}
\bibinfo{author}{Eberhardt, K.}, \bibinfo{author}{Stiebing, C.},
  \bibinfo{author}{Matth{\"a}us, C.}, \bibinfo{author}{Schmitt, M.} \&
  \bibinfo{author}{Popp, J.}
\newblock \bibinfo{title}{Advantages and limitations of raman spectroscopy for
  molecular diagnostics: an update}.
\newblock \emph{\bibinfo{journal}{Expert review of molecular diagnostics}}
  \textbf{\bibinfo{volume}{15}}, \bibinfo{pages}{773--787}
  (\bibinfo{year}{2015}).

\bibitem{chang2019antibiotic}
\bibinfo{author}{Chang, K.-W.} \emph{et~al.}
\newblock \bibinfo{title}{Antibiotic susceptibility test with surface-enhanced
  raman scattering in a microfluidic system}.
\newblock \emph{\bibinfo{journal}{Analytical chemistry}}
  \textbf{\bibinfo{volume}{91}}, \bibinfo{pages}{10988--10995}
  (\bibinfo{year}{2019}).

\bibitem{strola2014single}
\bibinfo{author}{Strola, S.~A.} \emph{et~al.}
\newblock \bibinfo{title}{Single bacteria identification by raman
  spectroscopy}.
\newblock \emph{\bibinfo{journal}{Journal of biomedical optics}}
  \textbf{\bibinfo{volume}{19}}, \bibinfo{pages}{111610}
  (\bibinfo{year}{2014}).

\bibitem{de2021biochemical}
\bibinfo{author}{de~Siqueira~e Oliveira, F.~S.}, \bibinfo{author}{da~Silva,
  A.~M.}, \bibinfo{author}{Pacheco, M. T.~T.}, \bibinfo{author}{Giana, H.~E.}
  \& \bibinfo{author}{Silveira, L.}
\newblock \bibinfo{title}{Biochemical characterization of pathogenic bacterial
  species using raman spectroscopy and discrimination model based on selected
  spectral features}.
\newblock \emph{\bibinfo{journal}{Lasers in Medical Science}}
  \textbf{\bibinfo{volume}{36}}, \bibinfo{pages}{289--302}
  (\bibinfo{year}{2021}).

\bibitem{Duraipandian2019}
\bibinfo{author}{Duraipandian, S.}, \bibinfo{author}{Petersen, J.} \&
  \bibinfo{author}{Lassen, M.}
\newblock \bibinfo{title}{Authenticity and concentration analysis of extra
  virgin olive oil using spontaneous raman spectroscopy and multivariate data
  analysis}.
\newblock \emph{\bibinfo{journal}{Appl. Sci.}} \textbf{\bibinfo{volume}{9}},
  \bibinfo{pages}{2433} (\bibinfo{year}{2019}).

\bibitem{garcia2018label}
\bibinfo{author}{Garc{\'\i}a-Timermans, C.} \emph{et~al.}
\newblock \bibinfo{title}{Label-free raman characterization of bacteria calls
  for standardized procedures}.
\newblock \emph{\bibinfo{journal}{Journal of microbiological methods}}
  \textbf{\bibinfo{volume}{151}}, \bibinfo{pages}{69--75}
  (\bibinfo{year}{2018}).

\bibitem{rousseau2021fast}
\bibinfo{author}{Rousseau, A.~N.} \emph{et~al.}
\newblock \bibinfo{title}{Fast antibiotic susceptibility testing via raman
  microspectrometry on single bacteria: An mrsa case study}.
\newblock \emph{\bibinfo{journal}{ACS omega}} \textbf{\bibinfo{volume}{6}},
  \bibinfo{pages}{16273--16279} (\bibinfo{year}{2021}).

\bibitem{kloss2015toward}
\bibinfo{author}{Klo\ss, S.}, \bibinfo{author}{Rösch, P.},
  \bibinfo{author}{Pfister, W.}, \bibinfo{author}{Kiehntopf, M.} \&
  \bibinfo{author}{Popp, J.}
\newblock \bibinfo{title}{Toward culture-free raman spectroscopic
  identification of pathogens in ascitic fluid}.
\newblock \emph{\bibinfo{journal}{Analytical chemistry}}
  \textbf{\bibinfo{volume}{87}}, \bibinfo{pages}{937--943}
  (\bibinfo{year}{2015}).

\bibitem{franco2019advances}
\bibinfo{author}{Franco-Duarte, R.} \emph{et~al.}
\newblock \bibinfo{title}{Advances in chemical and biological methods to
  identify microorganisms-from past to present. microorganisms}
  (\bibinfo{year}{2019}).

\bibitem{pahlow2015isolation}
\bibinfo{author}{Pahlow, S.} \emph{et~al.}
\newblock \bibinfo{title}{Isolation and identification of bacteria by means of
  raman spectroscopy}.
\newblock \emph{\bibinfo{journal}{Advanced drug delivery reviews}}
  \textbf{\bibinfo{volume}{89}}, \bibinfo{pages}{105--120}
  (\bibinfo{year}{2015}).

\bibitem{lussier2020deep}
\bibinfo{author}{Lussier, F.}, \bibinfo{author}{Thibault, V.},
  \bibinfo{author}{Charron, B.}, \bibinfo{author}{Wallace, G.~Q.} \&
  \bibinfo{author}{Masson, J.-F.}
\newblock \bibinfo{title}{Deep learning and artificial intelligence methods for
  raman and surface-enhanced raman scattering}.
\newblock \emph{\bibinfo{journal}{TrAC Trends in Analytical Chemistry}}
  \textbf{\bibinfo{volume}{124}}, \bibinfo{pages}{115796}
  (\bibinfo{year}{2020}).

\bibitem{lu2020combination}
\bibinfo{author}{Lu, W.}, \bibinfo{author}{Chen, X.}, \bibinfo{author}{Wang,
  L.}, \bibinfo{author}{Li, H.} \& \bibinfo{author}{Fu, Y.~V.}
\newblock \bibinfo{title}{Combination of an artificial intelligence approach
  and laser tweezers raman spectroscopy for microbial identification}.
\newblock \emph{\bibinfo{journal}{Analytical Chemistry}}
  \textbf{\bibinfo{volume}{92}}, \bibinfo{pages}{6288--6296}
  (\bibinfo{year}{2020}).

\bibitem{parmar2018image}
\bibinfo{author}{Parmar, N.} \emph{et~al.}
\newblock \bibinfo{title}{Image transformer} \bibinfo{pages}{4055--4064}
  (\bibinfo{year}{2018}).

\bibitem{shao2021adversarial}
\bibinfo{author}{Shao, R.}, \bibinfo{author}{Shi, Z.}, \bibinfo{author}{Yi,
  J.}, \bibinfo{author}{Chen, P.-Y.} \& \bibinfo{author}{Hsieh, C.-J.}
\newblock \bibinfo{title}{On the adversarial robustness of visual
  transformers}.
\newblock \emph{\bibinfo{journal}{arXiv e-prints}} \bibinfo{pages}{arXiv--2103}
  (\bibinfo{year}{2021}).

\bibitem{Hassani2021}
\bibinfo{author}{Hassani, A.} \emph{et~al.}
\newblock \bibinfo{title}{Escaping the big data paradigm with compact
  transformers}.
\newblock \emph{\bibinfo{journal}{arXiv preprint arXiv:2104.05704}}
  (\bibinfo{year}{2021}).
\newblock \urlprefix\url{https://arxiv.org/pdf/2104.05704.pdf}.

\bibitem{kong2015raman}
\bibinfo{author}{Kong, K.}, \bibinfo{author}{Kendall, C.},
  \bibinfo{author}{Stone, N.} \& \bibinfo{author}{Notingher, I.}
\newblock \bibinfo{title}{Raman spectroscopy for medical diagnostics—from
  in-vitro biofluid assays to in-vivo cancer detection}.
\newblock \emph{\bibinfo{journal}{Advanced drug delivery reviews}}
  \textbf{\bibinfo{volume}{89}}, \bibinfo{pages}{121--134}
  (\bibinfo{year}{2015}).

\bibitem{Zhang2010}
\bibinfo{author}{Zhang, Z.~M.}, \bibinfo{author}{Chen, S.} \&
  \bibinfo{author}{Liang, Y.~Z.}
\newblock \bibinfo{title}{Baseline correction using adaptive iteratively
  reweighted penalized least squares}.
\newblock \emph{\bibinfo{journal}{Analyst}} \textbf{\bibinfo{volume}{135}},
  \bibinfo{pages}{1138--1146} (\bibinfo{year}{2010}).

\end{thebibliography}

\clearpage

%=====================================================

\appendix
%\FloatBarrier

\setcounter{page}{1}
\renewcommand{\thepage}{SI~\arabic{page}}

\setcounter{figure}{0}
\renewcommand{\thefigure}{SI\arabic{figure}}

\setcounter{table}{0}
\renewcommand{\thetable}{SI\arabic{table}}

\setcounter{equation}{0}
\renewcommand{\theequation}{SI~\thesection.\arabic{equation}}

\onecolumngrid

\clearpage

\section*{Acknowledgments}
We gratefully acknowledge fruitful conversations with Poul A. Jessen from BacAlert. This research was funded by the Danish Agency for Institutions and Educational Grants and the Innovation Fund Denmark (IFD) under Eurostars project Bacsens (case No. 9046-00032A).

\section*{Author Contributions}
J.B.C., I.U. and M.L. designed and built the Raman microscope. R.B.G. and T.E.A. prepared the bacteria for the experiments. J.B.C, O.R., and M.L obtained the main experimental Raman results. B.L.T. designed and developed the software for machine learning analysis. The paper was written by B.L.T., J.B.C. and M.L. with contributions from all authors. M.L. conceived and supervised the research. 

\section*{Competing interests.}
The authors declare no competing financial interests. 

\section*{Data Availability Statement}
The data that support the findings of this study are available from the corresponding author upon reasonable request.

\section*{Correspondence}
Correspondence and requests for materials should be addressed to ML (ml@dfm.dk).

\clearpage

\end{document}